\documentclass[conference]{IEEEtran}
\IEEEoverridecommandlockouts
\usepackage{cite}
\usepackage{amsmath,amssymb,amsfonts}
\usepackage{algorithmic}
\usepackage{graphicx}
\usepackage{textcomp}
\usepackage{xcolor}
\usepackage{amssymb}
\usepackage{tikz}
\usepackage{array}
\usepackage{comment}
\usepackage{hyperref}
\usetikzlibrary{shapes,arrows}
\usetikzlibrary{positioning, arrows.meta}

\usepackage{booktabs}

\def\BibTeX{{\rm B\kern-.05em{\sc i\kern-.025em b}\kern-.08em
    T\kern-.1667em\lower.7ex\hbox{E}\kern-.125emX}}
\begin{document}

\title{Arabic Handwritten Text for Person Biometric Identification: A Deep Learning Approach}

\author{\IEEEauthorblockN{1\textsuperscript{st} Mazen Balat}
\IEEEauthorblockA{\textit{Computer science and information technology} \\
\textit{Egypt-Japan University of Science and Technology (E-JUST)}\\
Alexandria, 21934, Egypt \\
mazen.balat@ejust.edu.eg}
\and
\IEEEauthorblockN{1\textsuperscript{st} Youssef Mohamed}
\IEEEauthorblockA{\textit{Computer science and information technology} \\
\textit{Egypt-Japan University of Science and Technology (E-JUST)}\\
Alexandria, 21934, Egypt \\
youssef.khalil@ejust.edu.eg}
\and
\IEEEauthorblockN{2\textsuperscript{nd} Ahmed Heakl}
\IEEEauthorblockA{\textit{Computer engineering} \\
\textit{Egypt-Japan University of Science and Technology (E-JUST)}\\
Alexandria, 21934, Egypt \\
ahmed.heakl@ejust.edu.eg}
\and
\IEEEauthorblockN{3\textsuperscript{rd} Ahmed B. Zaky}
\IEEEauthorblockA{\textit{Computer science and information technology} \\
\textit{Egypt-Japan University of Science and Technology (E-JUST)}\\
Alexandria, 21934, Egypt \\
ahmed.zaky@feng.bu.edu.eg,ahmed.zaky@ejust.edu.eg}
}

\newcommand{\heakl}[1]{\textcolor{blue}{#1}}

\maketitle

\begin{abstract}
This study thoroughly investigates how well deep learning models can recognize Arabic handwritten text for person biometric identification. It compares three advanced architectures—ResNet50, MobileNetV2, and EfficientNetB7—using three widely recognized datasets: AHAWP, Khatt, and LAMIS-MSHD. Results show that EfficientNetB7 outperforms the others, achieving test accuracies of 98.57\%, 99.15\%, and 99.79\% on AHAWP, Khatt, and LAMIS-MSHD datasets, respectively. EfficientNetB7's exceptional performance is credited to its innovative techniques, including compound scaling, depth-wise separable convolutions, and squeeze-and-excitation blocks. These features allow the model to extract more abstract and distinctive features from handwritten text images. The study's findings hold significant implications for enhancing identity verification and authentication systems, highlighting the potential of deep learning in Arabic handwritten text recognition for person biometric identification.

\end{abstract}


\section{Introduction}

Biometric identification has become a crucial aspect of modern security and authentication systems. Among various biometric modalities, handwritten text has gained significant attention due to its unique characteristics and potential applications. In this work, we focus on Arabic handwriting recognition, which presents a challenging task due to the cursive nature of the script and limited availability of datasets.

Recent advances in deep learning have improved the accuracy of biometric identification systems, including fingerprint\cite{10255275}, face\cite{Liu2023}, iris\cite{yin2023deep}, and voice recognition\cite{Saritha2024},electroencephalography (EEG)\cite{10217750}. However, handwritten text recognition remains a relatively understudied area, particularly for Arabic scripts. Our research aims to bridge this gap by exploring the potential of Arabic handwriting recognition for biometric identification.

In this paper, we present a novel approach to Arabic handwriting recognition, leveraging advanced preprocessing and augmentation techniques to improve model accuracy and robustness. We also investigate the effectiveness of transfer learning in reducing training time and enhancing recognition accuracy. Our experimental results demonstrate the feasibility of Arabic handwriting recognition for biometric identification and its potential applications in forensic document examination, identity verification, and access control.

The main contributions of this work are:
\begin{enumerate}
\item Introduced new preprocessing and augmentation techniques for Arabic handwriting recognition, improving model accuracy and robustness.
\item Demonstrated the potential of transfer learning to improve Arabic handwriting recognition accuracy and reduce training time.
\item Conducted a study on the relationship between the number of writers and system accuracy.
\end{enumerate}

Our approach has the potential to be extended to other languages and handwriting styles, making it a versatile method for biometric identification.

\section{Related Works}

In "GR-RNN: Global-Context Residual Recurrent Neural Networks for Writer Identification"\cite{he2021grrnn},the proposed methodology involves a series of convolutional layers, max-pooling layers, and global average pooling, followed by the integration of extracted features using a residual RNN to model spatial dependencies between fragments. The approach is extensively evaluated on four benchmark datasets, comprising IAM (657 writers), CVL (310 writers), Firemaker (250 writers), and CERUG-EN (105 writers). The results demonstrate the efficacy of the proposed GR-RNN method, achieving top-1 accuracy ranging from 82.4\% to 95.2\% for word-level writer identification, 82.4\% to 95.2\% for line-level writer identification, and 82.6\% to 96.6\% for page-level writer identification. These findings underscore the effectiveness of the GR-RNN approach in capturing the global context and spatial dependencies inherent in handwritten images, thereby facilitating accurate writer identification.

In "Hybrid Trainable System for Writer Identification of Arabic Handwriting"\cite{hybird},the writer identification scheme is evaluated on the KHATT dataset, comprising 4,000 Arabic handwritten documents from 1,000 authors. The cumulative match characteristic (CMC) curve is used to measure the performance of the identification algorithm. Experimental results show that the proposed hybrid trainable system, combining convolutional neural networks (CNNs) and support vector machines (SVMs), outperforms other models, including HOG and ANN, and ResNet50. The system achieves an accuracy of 94.2\% when using the whole paragraph with augmentation, and 83.2\% when using lines. The results demonstrate the effectiveness of the proposed system in writer identification tasks, and highlight the importance of using augmentation to increase the training dataset. Future work will focus on identifying effective features that can facilitate writer identification based on text lines and subwords with high accuracy.

In ”Writer identification using textural features”\cite{lamisidentification}, the authors proposed a writer identification approach using local binary pattern (LBP) features and vector of locally aggregated descriptors (VLAD) encoding. They evaluated their method on the LAMIS-MSHD dataset, which consists of Arabic and French handwritten texts from 100 individuals. The experimental results showed that the proposed approach achieved high writer identification rates, with top-1 classification rates ranging from 95\% to 100\% for both Arabic and French texts, depending on the number of clusters used in the VLAD encoding. Notably, the authors observed that medium and small numbers of clusters (between 16 and 256) resulted in the highest identification rates, while larger numbers of clusters led to degraded performance. This study demonstrates the effectiveness of texture-based features and VLAD encoding for writer identification in handwritten Arabic and French texts.

In "Offline text-independent writer identification using a codebook with structural features"\cite{10.1371/journal.pone.0284680}, the authors proposed a writer identification approach using contour-based features extracted from handwritten texts. They employed two feature extraction methods, CPCA and CON3, and encoded the features into a codebook using k-means clustering. The authors evaluated their approach on two datasets, KHATT and IAM, and achieved high identification rates of up to 96.3\% and 88.2\%, respectively, using multiclass SVM and nearest neighbor classifiers. The authors analyzed the impact of various parameters, including contour fragment length and angular quantization, on the system's performance and found that they have a significant effect. The study demonstrates the effectiveness of using structural features and codebook-based encoding for offline text-independent writer identification.

In the realm of biometric identification, the efficacy of deep learning models is paramount, particularly in the challenging domain of Arabic handwritten text recognition. This study harnesses the robust capabilities of transfer learning, employing state-of-the-art convolutional neural networks such as ResNet50, MobileNetV2, and EfficientNetB7. These models, pre-trained on extensive image datasets, have demonstrated exceptional adaptability in feature extraction and pattern recognition, making them ideal candidates for the task at hand. ResNet50\cite{he2015deep}, with its deep residual learning framework, facilitates the training of networks that are substantially deeper than those used previously. MobileNetV2\cite{sandler2019mobilenetv2}, designed for mobile and embedded vision applications, offers an optimal balance between latency and accuracy. EfficientNetB7\cite{tan2020efficientnet}, the largest variant in the EfficientNet family, achieves superior performance by scaling up the network in a more structured manner. The integration of these models into our framework, through transfer learning, has significantly advanced the frontiers of Arabic handwritten text recognition, providing a testament to their versatility and power in deciphering complex linguistic patterns inherent to the Arabic script.

\section{Datasets}

\subsection{KHATT Dataset}

\begin{figure}[h]
\centering
\includegraphics[width=0.4\textwidth]{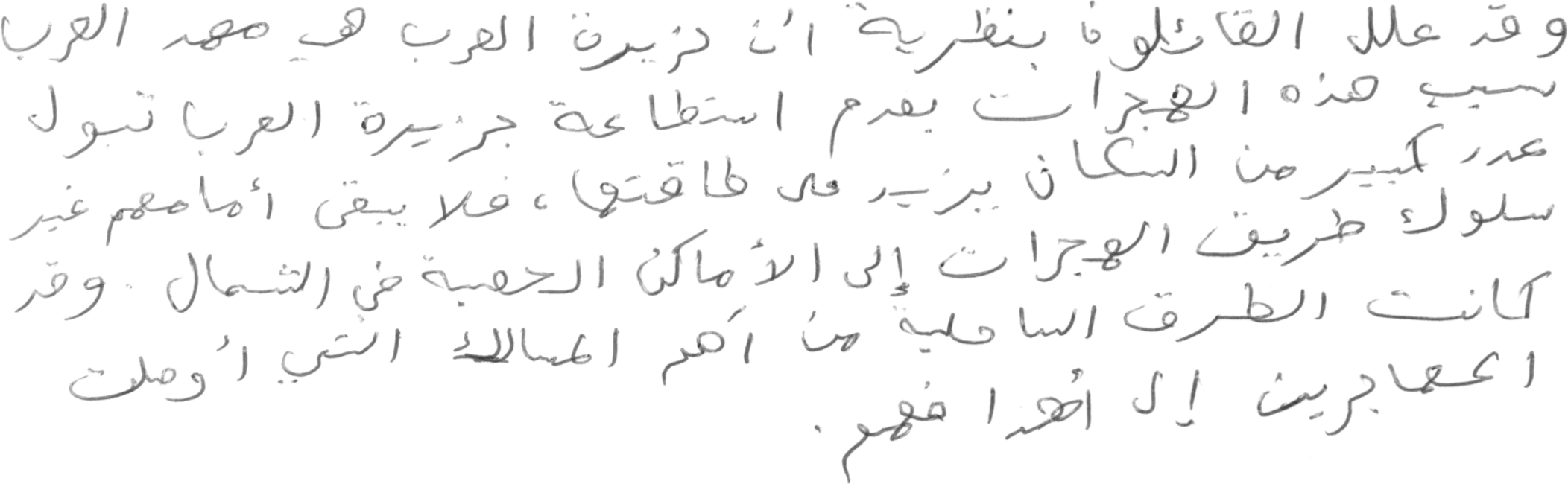}
\caption{KHATT dataset sample}
\label{fig:khatt}
\end{figure}

We use the KHATT database\cite{Khatt}, a comprehensive collection of handwritten Arabic text from 1,000 writers, with 2,000 unique paragraph images and 2,000 similar-text paragraph images (see Figure \ref{fig:khatt}).

\subsection{AHAWP Dataset}

\begin{figure}[h]
\centering
\includegraphics[width=0.4\textwidth]{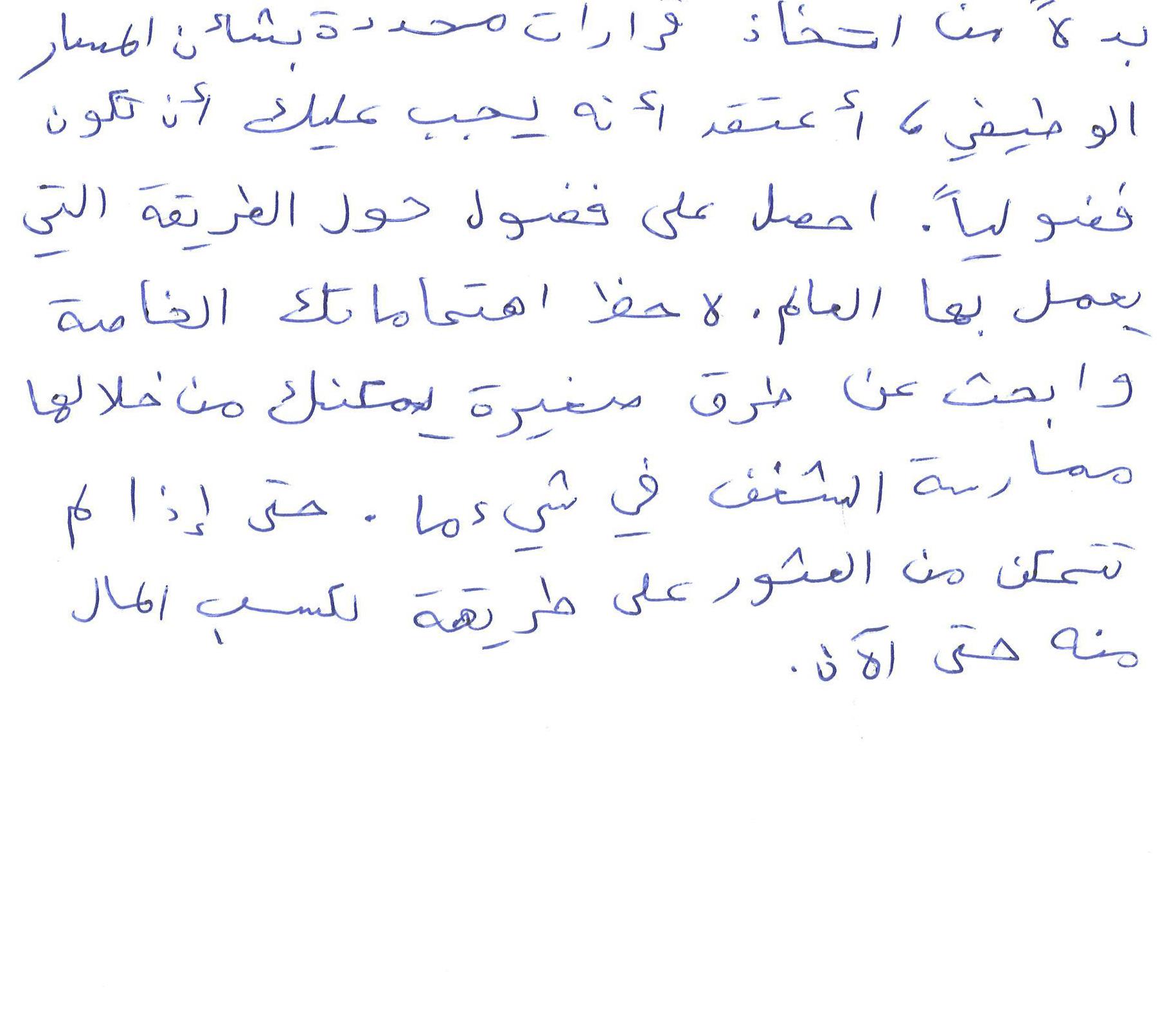}
\caption{AHAWP dataset sample}
\label{fig:ahawp}
\end{figure}

The AHAWP dataset\cite{Khan2022} provides a diverse collection of handwritten Arabic alphabets, words, and paragraphs from 82 individuals, with 53,199 alphabet images, 8,144 word images, and 241 paragraph images (see Figure \ref{fig:ahawp}).

\subsection{LAMIS-MSHD Dataset}

\begin{figure}[h]
\centering
\includegraphics[width=0.4\textwidth]{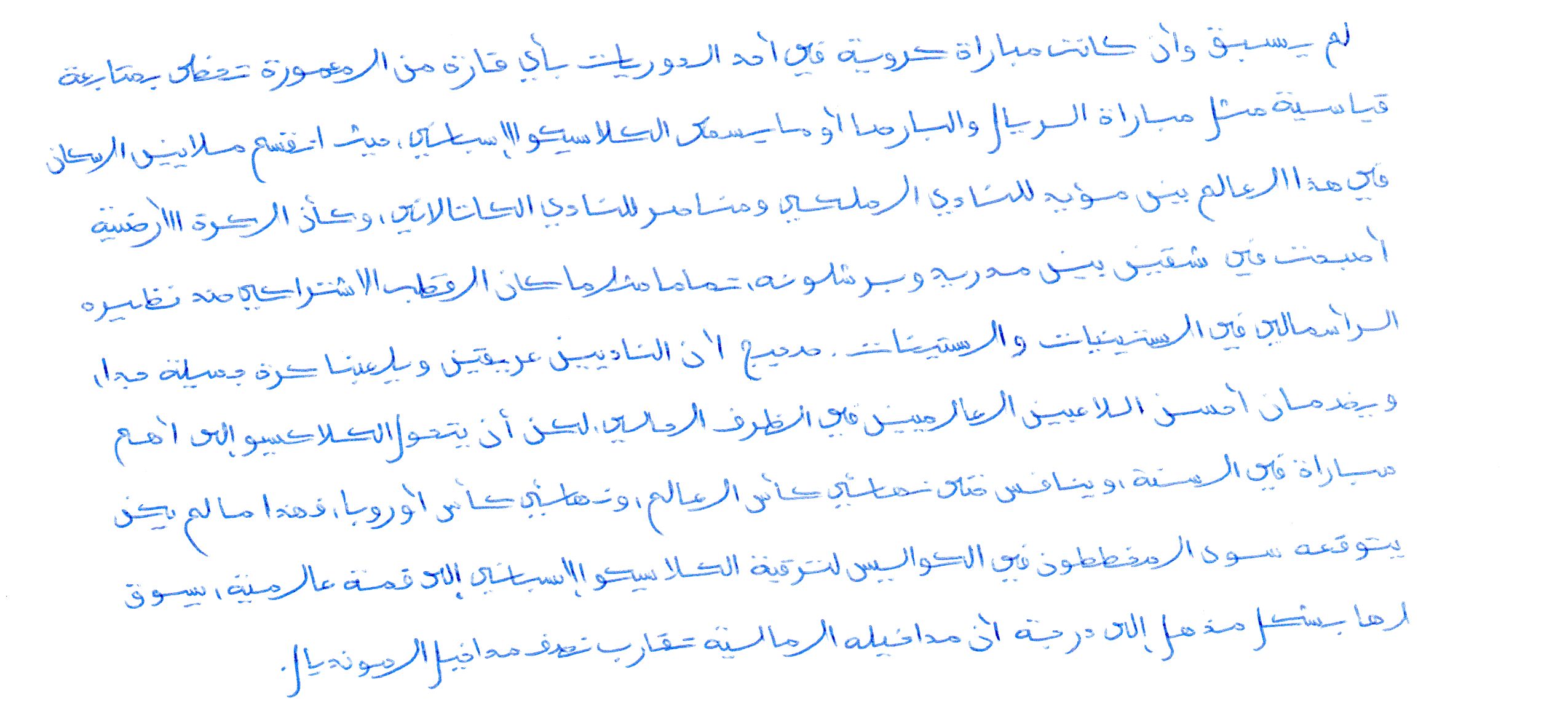}
\caption{LAMIS-MSHD dataset sample}
\label{fig:lamis}
\end{figure}

The LAMIS-MSHD dataset\cite{djeddi2015lamis} is a multi-script offline handwritten database containing Arabic and French text samples, signatures, and digits from 100 writers (see Figure \ref{fig:lamis}).

\begin{table}[htbp]
\caption{Comparison of Datasets}
\label{tab:datasets} 
\begin{center}
\begin{tabular}{|p{2cm}|p{1.2cm}|p{1.2cm}|p{1.2cm}|p{1.2cm}|} 
\hline
\textbf{Dataset} & \textbf{Number of Writers} & \textbf{Paragraphs per User} & \textbf{Total Paragraphs} & \textbf{Words per User} \\
\hline
KHATT & 1,000 & 4 & 4,000 & -  \\
AHAWP & 82 & 3 & 246 & 100 \\
LAMIS-MSHD & 100 & 12 & 1,200 & -  \\
\hline
\end{tabular}
\end{center}
\end{table}

\section{Preprocessing}

To advance person biometric identification using deep learning models for Arabic handwritten text recognition, a crucial preprocessing step is necessary to prepare the data for training and testing. Our proposed preprocessing pipeline consists of five stages: binarization, dilation, contour detection and sorting, filtering small components, and ROI extraction.

\subsection{Binarization}
\begin{figure}[htbp]
    \centering
    \includegraphics[width=0.4\textwidth]{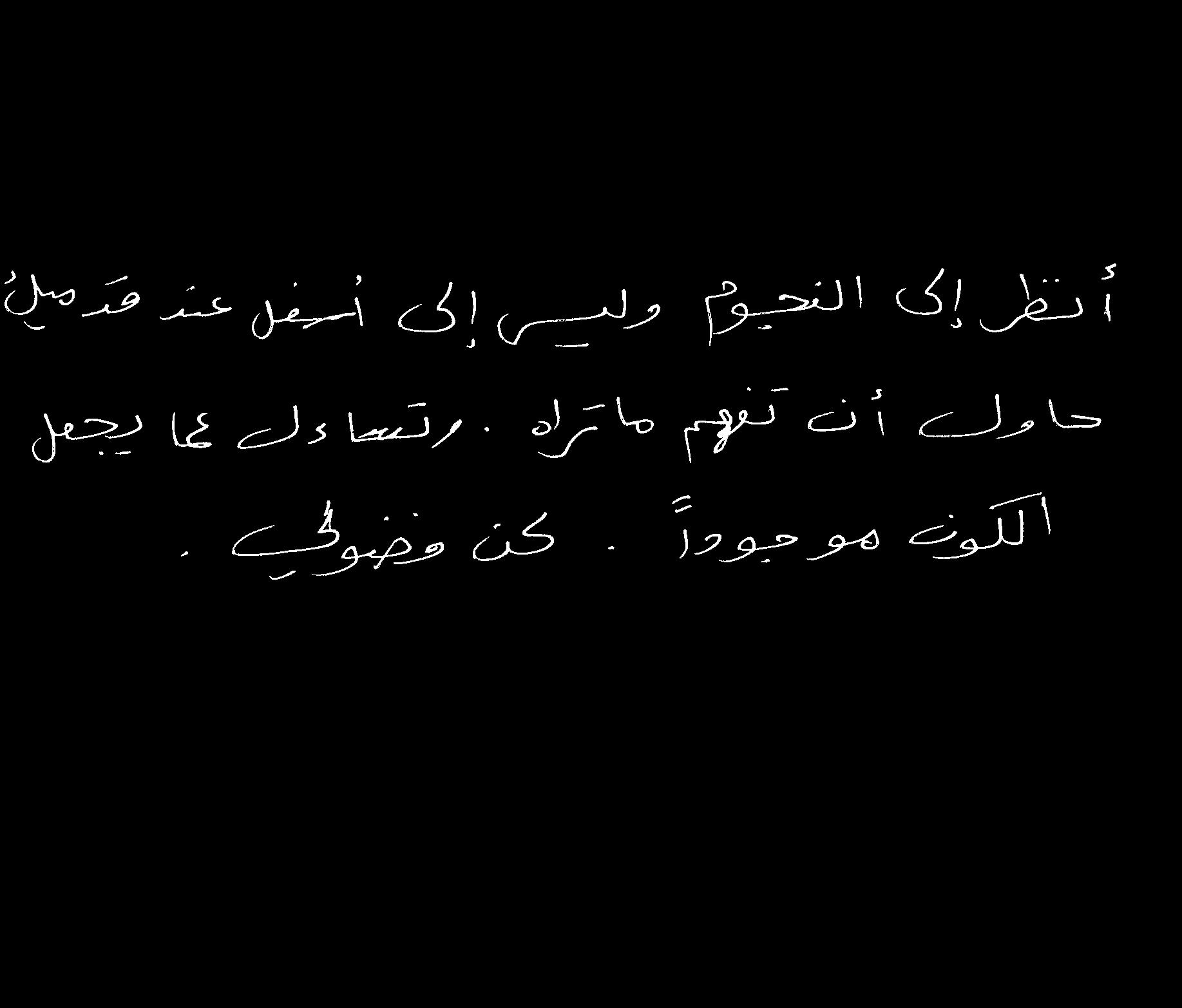}
    \caption{Binarization result}
    \label{fig:binarization}
\end{figure}

The first step in our preprocessing pipeline is binarization\cite{Otsu1979ATS}, which involves converting the grayscale images of Arabic handwritten paragraphs into binary images. This is done to enhance the contrast between the text and the background, making it easier to extract features from the text. We applied the Otsu's thresholding method to achieve binarization (see Figure \ref{fig:binarization}).

\subsection{Dilation}
\begin{figure}[htbp]
    \centering
    \includegraphics[width=0.4\textwidth]{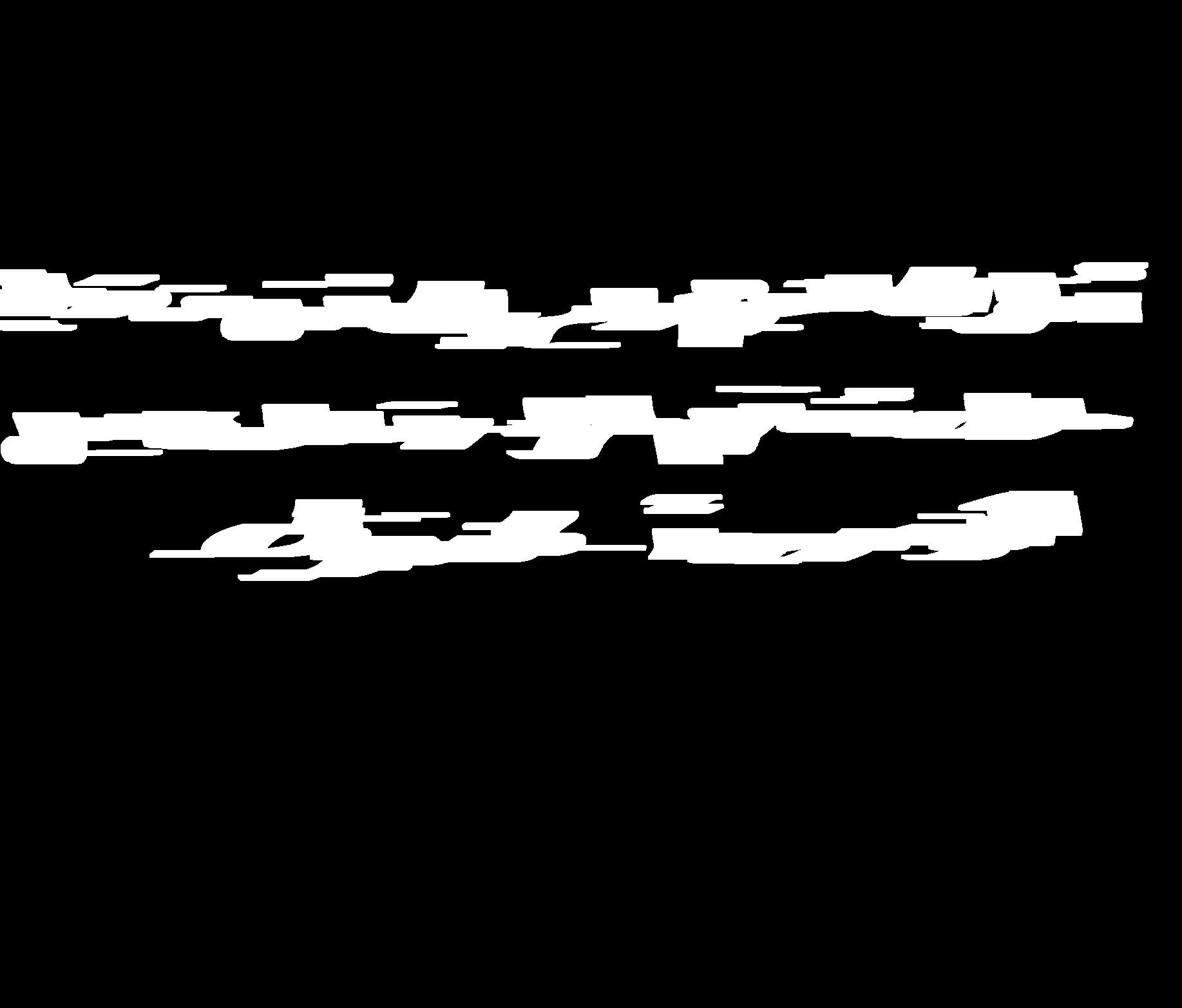}
    \caption{Dilation result}
    \label{fig:dilation}
\end{figure}

After binarization, we applied dilation\cite{dilation} to the binary images to fill in any gaps between connected components. This step helps to strengthen the connections between adjacent strokes in the Arabic handwritten text, making it easier to detect contours (see Figure \ref{fig:dilation}).

\subsection{Contour Detection and Sorting}
\begin{figure}[htbp]
    \centering
    \includegraphics[width=0.4\textwidth]{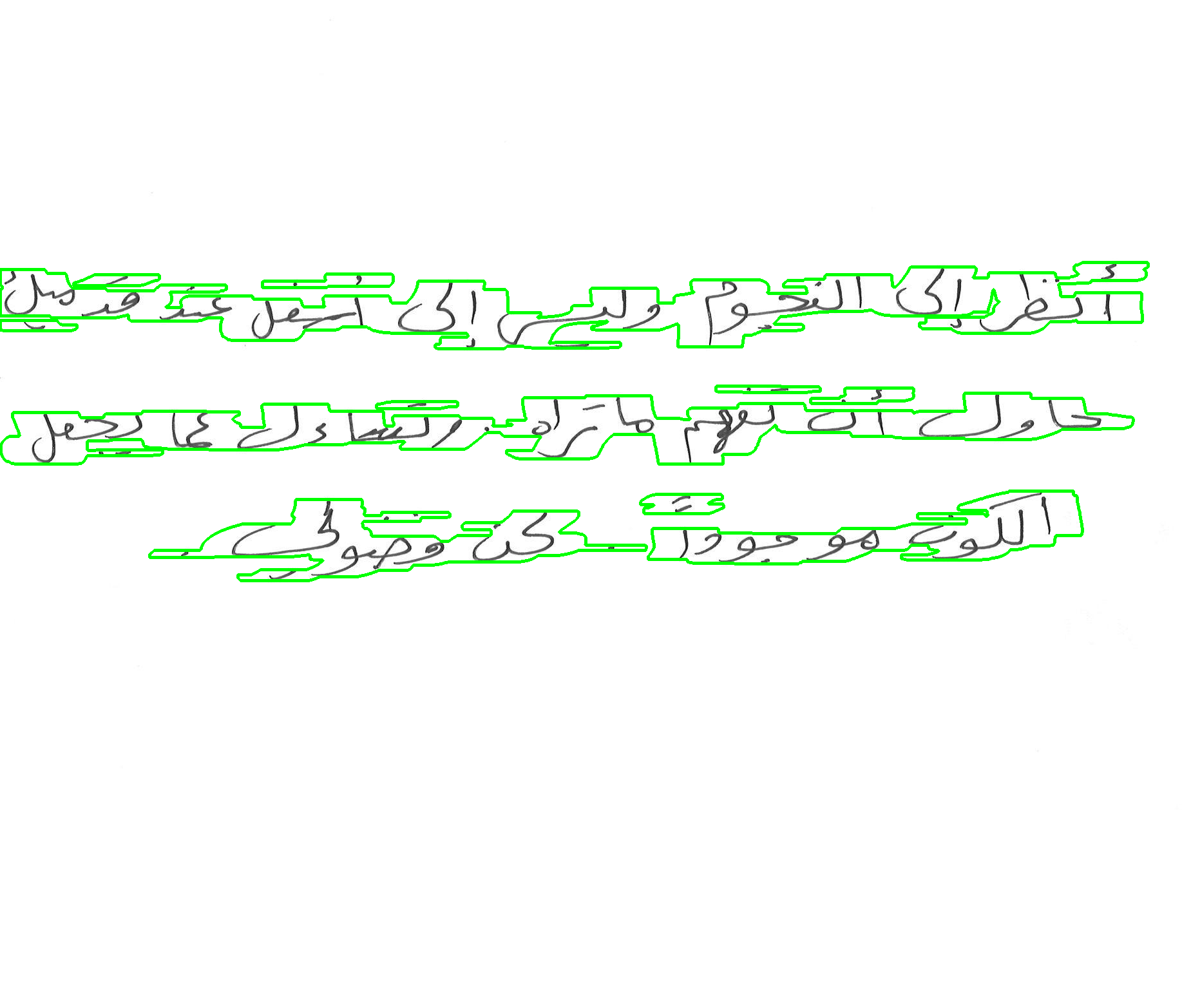}
    \caption{Contour Detection and Sorting result}
    \label{fig:contour_sorting}
\end{figure}

In the third stage, we detected contours in the dilated images using the Canny edge detection algorithm\cite{4767851}. The detected contours were then sorted based on their size and orientation to identify the text lines. This step is crucial in separating the text into individual lines, which is essential for Arabic handwritten text recognition (see Figure \ref{fig:contour_sorting}).

\subsection{ROI Extraction with Filtering}
\begin{figure}[htbp]
    \centering
    \includegraphics[width=0.7\linewidth]{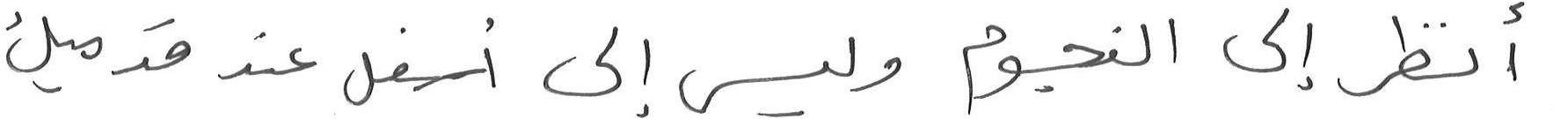}
    \vspace{10pt} % Adjust the amount of space as needed
    
    \includegraphics[width=0.7\linewidth]{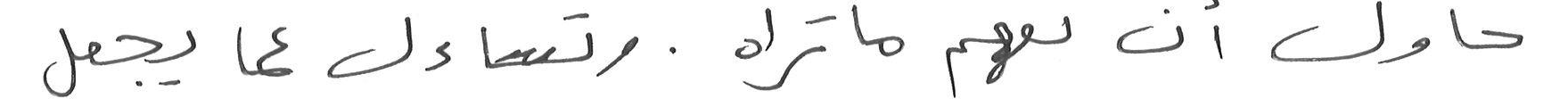}
    \vspace{10pt} % Adjust the amount of space as needed
    
    \includegraphics[width=0.7\linewidth]{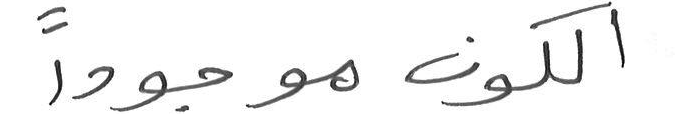}
    \vspace{10pt} % Adjust the amount of space as needed
    
    \includegraphics[width=0.7\linewidth]{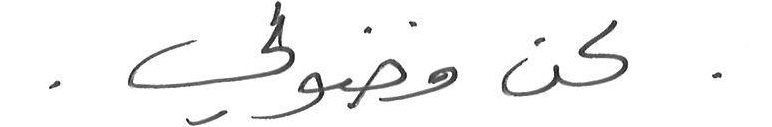}
    \caption{ROI Extraction with Filtering result}
    \label{fig:roi_extraction}
\end{figure}

Finally, we extracted the regions of interest (ROIs)\cite{Roi} from the filtered images, which correspond to individual text lines. Each ROI represents a single line of Arabic handwritten text, which is then fed into the deep learning model for recognition. During ROI extraction, we also filter out small components that are unlikely to be part of the text (see Figure \ref{fig:roi_extraction}).

\medskip

Finally, each extracted ROI is resized to a fixed size of 224x224 pixels to ensure uniformity for the deep learning model.

\subsection{Data Splitting}
After preprocessing all the images, the data is split into training, validation, and testing sets. In this work, we adopt an 80/10/10 split, where 80\% of the data is used for training the deep learning model, 10\% is used for validation, and the remaining 10\% is used for testing. The validation set allows for tuning the hyperparameters of the model, while the testing set provides an unbiased evaluation of the model's generalization performance on unseen data.

By adopting a rigorous data splitting approach, we can ensure that our model is thoroughly trained, validated, and tested, thereby enhancing its reliability and effectiveness in addressing complex problems.

\section{Data Augmentation}

To increase the robustness of our deep learning models for Arabic handwritten text recognition, we employed several data augmentation techniques. These techniques aim to artificially increase the size of the training dataset by applying transformations to the original images.

\subsection{Reduce Line Thickness}

\begin{figure}[ht]
    \centering
    \includegraphics[width=0.5\textwidth]{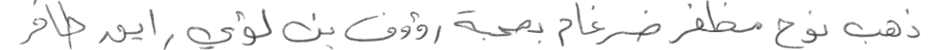}
    \includegraphics[width=0.5\textwidth]{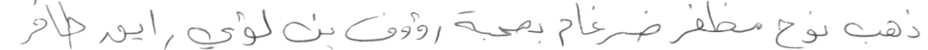}
    \caption{Original image and after applying the reduce line thickness technique.}
    \label{fig:line_thickness}
\end{figure}

One of the augmentation techniques used is reducing the line thickness of the handwritten text. This technique helps to simulate the variability in writing styles and instruments. Figure \ref{fig:line_thickness} illustrates the effect of this technique on a sample image.

\subsection{Apply Random Noise}

\begin{figure}[ht]
    \centering
    \includegraphics[width=0.5\textwidth]{imgs/orignalbefore.png}
    \includegraphics[width=0.5\textwidth]{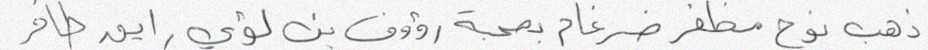}
    \caption{Original image and after applying the random noise technique.}
    \label{fig:random_noise}
\end{figure}

Another technique used is adding random noise to the handwritten text. This technique helps to simulate the noise and distortions that may occur during image acquisition or transmission. Figure \ref{fig:random_noise} shows the effect of this technique on a sample image.

\subsection{Apply Random Stretch}

\begin{figure}[ht]
    \centering
    \includegraphics[width=0.5\textwidth]{imgs/orignalbefore.png}
    \includegraphics[width=0.5\textwidth]{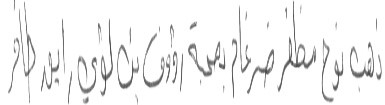}
    \caption{Original image and after applying the random stretch technique.}
    \label{fig:random_stretch}
\end{figure}

We also applied random stretch to the handwritten text, horizontally stretching the image by a random factor between -0.9 and +0.1. This technique helps to simulate the variability in writing sizes and orientations. Figure \ref{fig:random_stretch} illustrates the effect of this technique on a sample image.

\subsection{Combined Augmentations (All Augmentations)}

All three augmentation techniques were combined to further increase the diversity of the training dataset.

The ratio of the final dataset size to the original dataset size after applying each augmentation technique and their combinations are summarized below:

\begin{center}
    \begin{tabular}{|l|c|}
        \hline
        \textbf{Augmentation Technique} & \textbf{Final Dataset Size Ratio} \\
        \hline
        Reduce Line Thickness & 2x \\
        Apply Random Noise & 3x \\
        Apply Random Stretch & 4x \\
        \hline
    \end{tabular}
\end{center}

These data augmentation techniques effectively quadrupled the dataset size when all techniques were combined. This increased diversity in the training data is expected to enhance the performance and robustness of our deep learning models across various datasets for Arabic handwritten text recognition.

\section{Methodology}

In this study, we leveraged the power of transfer learning to fine-tune three state-of-the-art deep learning models - ResNet50, MobileNetV2, and EfficientNetB7 - for Arabic handwritten text recognition in the context of person biometric identification. By building upon the knowledge these models have already acquired, we aimed to create a robust and accurate recognition system that can effectively identify individuals.

\subsection{Model Fine-Tuning}

To adapt these pre-trained models to our specific task, we employed a fine-tuning approach. This involved modifying the final classification layer of each model to accommodate the number of classes in our dataset, allowing the models to learn features that are specific to our Arabic handwritten text dataset.

\begin{figure}
    \centering
    \begin{tikzpicture}[node distance=1cm]
        % Nodes
        \node[draw, minimum width=3cm, minimum height=1cm] (pretrained) {Pre-trained Model};
        \node[draw, below=of pretrained, minimum width=3cm, minimum height=1cm] (featuremaps) {Feature Maps};
        \node[draw, below=of featuremaps, minimum width=3cm, minimum height=1cm] (fcnew) {FC New};
        \node[draw, below=of fcnew, minimum width=3cm, minimum height=1cm] (output) {Output};
        
        % Arrows
        \draw[-Latex] (pretrained) -- (featuremaps);
        \draw[-Latex, dashed] (featuremaps) -- (fcnew);
        \draw[-Latex] (fcnew) -- (output);
        
        % Labels
        \node[left=of pretrained] (input) {Input};
        \node[right=of output] (predictions) {Predictions};
        
        % Arrows for labels
        \draw[-Latex] (input) -- (pretrained);
        \draw[-Latex] (output) -- (predictions);
    \end{tikzpicture}
    \caption{Illustration of Fine-Tuning Process}
    \label{fig:fine-tuning}
\end{figure}
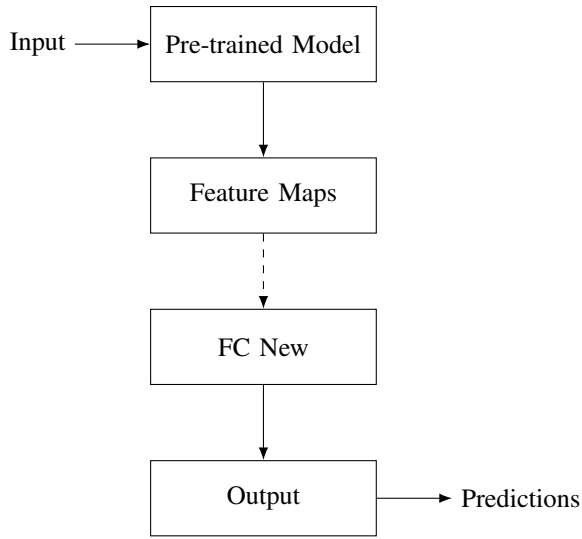

As shown in Figure \ref{fig:fine-tuning}, we replaced the last fully connected layer of each model with a new one that matches the number of classes in our dataset. This enabled the models to produce predictions that are relevant to our task, rather than the generic predictions they were initially trained to make.

\subsection{Optimizing Model Performance}

To optimize the performance of each model, we conducted a thorough training and hyperparameter tuning process. We trained each model using our augmented dataset, with a batch size of 16 and a maximum of 10 epochs. We employed the Adam optimizer with an initial learning rate of 0.0001, which was decayed by a factor of 0.1 every 7 epochs using a step learning rate scheduler.

\section{Results}

The initial experiments revealed the presence of overfitting across all datasets. To address this challenge, we implemented various data augmentation techniques, including adjusting line thickness, adding random noise, and randomly stretching the images. These techniques introduced diversity into the training data, which in turn facilitated better generalization. The combination of these augmentation methods significantly reduced overfitting, resulting in improved performance across all datasets.

\subsection{AHAWP Dataset}

\begin{table}
  \centering
  \renewcommand{\arraystretch}{1.5} % increase row height
  \begin{tabular}{lccc}
    \hline
    \multicolumn{4}{c}{Without Augmentation} \\
    \hline
    Model & Train Acc & Val Acc & Test Acc \\
    \hline
    ResNet &  \textbf{100.00\%} &  \textbf{87.50\%} &  \textbf{87.84\%} \\
    MobileNet & 99.00\% & 82.81\% & 82.40\% \\
    EfficientNet B7 & 98.74\% & 75.00\% & 75.00\% \\
    \hline
    \multicolumn{4}{c}{All Augmentations} \\
    \hline
    Model & Train Acc & Val Acc & Test Acc \\
    \hline
    ResNet & 99.35\% & 90.95\% & 90.82\% \\
    MobileNet & 99.88\% & 94.82\% & 94.4\% \\
    EfficientNet B7 &  \textbf{99.9\%} &  \textbf{98.57\%} &  \textbf{98.52\%} \\
    \hline
  \end{tabular}
  \caption{Comparison of models with different augmentations for the AHAWP dataset}
  \label{tab:AHAWB-model-comparison}
\end{table}

EfficientNet B7 consistently outperformed other models, exhibiting the highest test accuracy across all augmentation techniques (see Table \ref{tab:AHAWB-model-comparison}). Upon the application of all augmentations, a substantial improvement in performance was observed across all models. Notably, ResNet's test accuracy improved by 3.42\%, MobileNet's by 14.56\%, and EfficientNet B7's by an impressive 31.36\%. These findings underscore the pivotal role of augmentations in bolstering the efficacy of image classification models, with EfficientNet B7 showcasing the most significant enhancement.

\subsection{KHATT Dataset}

\begin{table}[htbp]
  \centering
  \renewcommand{\arraystretch}{1.5} % increase row height
  \begin{tabular}{lccc}
    \hline
    \multicolumn{4}{c}{No Augmentation} \\
    \hline
    Model & Train Acc & Test Acc & Val Acc \\
    \hline
    ResNet &  \textbf{94.48\%} & 31.99\% & 29.64\% \\
    MobileNet & 76.28\% &  \textbf{37.94\%} &  \textbf{36.58\%} \\
    EfficientNet B7 & 91.77\% & 32.45\% & 31.90\% \\
    \hline
    \multicolumn{4}{c}{All Augmentations} \\
    \hline
    ResNet &  \textbf{99.84\%} & 98.47\% & 98.71\% \\
    MobileNet & 98.76\% & 97.08\% & 97.28\% \\
    EfficientNet B7 & 99.76\% &  \textbf{99.15\%} &  \textbf{99.12\%} \\
    \hline
  \end{tabular}
  \caption{Comparison of models with different augmentations for the KHATT dataset}
  \label{tab:model-comparison-khatt}
\end{table}

EfficientNet B7 emerged as the preeminent model across all augmentation scenarios, boasting the highest test accuracy and validation accuracy (see Table \ref{tab:model-comparison-khatt}). Following augmentation, all models exhibited substantial performance enhancements. Specifically, ResNet's test accuracy surged by 66.48\%, MobileNet's by 59.14\%, and EfficientNet B7's by an impressive 66.70\%. These results underscore the efficacy of augmentations in fortifying the performance of image classification models, with all models demonstrating substantial progress.

Our model has demonstrated a significant enhancement, achieving an improvement of approximately 4.95\% over the accuracy reported in the study referenced as \cite{hybird}. Furthermore, it surpassed the accuracy reported in the paper cited as \cite{10.1371/journal.pone.0284680} by approximately 2.85\%.

.

\subsection{LAMIS-MSHD Dataset}

\begin{table}[htbp]
  \centering
  \renewcommand{\arraystretch}{1.5} % increase row height
  \begin{tabular}{lccc}
    \hline
    \multicolumn{4}{c}{No Augmentation} \\
    \hline
    Model & Train Acc & Test Acc & Val Acc \\
    \hline
    ResNet &  \textbf{99.71\%} & 79.96\% & \textbf{84.10\%} \\
    MobileNet & 98.93\% & 75.16\% & 75.10\% \\
    EfficientNet B7 & 99.52\% &  \textbf{83.20\%} & 82.15\% \\
    \hline
    \multicolumn{4}{c}{All Augmentations} \\
    \hline
    Model & Train Acc & Test Acc & Val Acc \\
    \hline
    ResNet & \textbf{99.89\%} & 99.37\% & 99.22\% \\
    MobileNet & 99.80\% & 98.48\% & 99.37\% \\
    EfficientNet B7 & 99.52\% & \textbf{99.79\%} & \textbf{99.63\%} \\
    \hline
  \end{tabular}
  \caption{Comparison of models with different augmentations for the LAMIS dataset}
  \label{tab:model-comparison-lamis}
\end{table}

EfficientNet B7 emerged as the top-performing model for the LAMIS dataset, maintaining the highest test accuracy both before and after augmentations (see Table \ref{tab:model-comparison-lamis}). Following augmentation, ResNet's test accuracy improved by 19.41\%, MobileNet's by 23.32\%, and EfficientNet B7's by 17.48\%. The consistent high performance of EfficientNet B7 across all datasets underscores its robustness and effectiveness in handwritten text recognition tasks.

\medskip

\section{Conclusion}

In this work, we present a comprehensive approach to Arabic handwritten text recognition for person biometric identification using deep learning. We address the challenge of accurately recognizing Arabic handwritten text, which is crucial for applications such as forensic document examination, identity verification, and access control. To solve this problem, we develop a robust preprocessing pipeline and employ data augmentation techniques to improve the accuracy and robustness of the recognition process. The combination of these techniques plays a crucial role in achieving significant improvements in recognition accuracy.

We also investigate the effectiveness of transfer learning, reducing training time and achieving higher recognition accuracy of up to \textbf{99.79\%}. Furthermore, our study explores the relationship between the number of writers and the system's accuracy, providing insights into the scalability of the proposed approach.

The results demonstrate the feasibility and potential of using Arabic handwritten text for person biometric identification, with our techniques leading to significant improvements in recognition accuracy. These findings pave the way for further advancements in Arabic handwriting recognition and its applications in biometric identification, and the proposed method can be extended to other languages and handwriting styles, making it a versatile solution for security and authentication systems.

\bibliographystyle{elsarticle-num} 
\bibliography{paper}

\begin{thebibliography}{10}
\expandafter\ifx\csname url\endcsname\relax
  \def\url#1{\texttt{#1}}\fi
\expandafter\ifx\csname urlprefix\endcsname\relax\def\urlprefix{URL }\fi
\expandafter\ifx\csname href\endcsname\relax
  \def\href#1#2{#2} \def\path#1{#1}\fi

\bibitem{10255275}
S.~A. Grosz, A.~K. Jain, Afr-net: Attention-driven fingerprint recognition network, IEEE Transactions on Biometrics, Behavior, and Identity Science 6~(1) (2024) 30--42.
\newblock \href {https://doi.org/10.1109/TBIOM.2023.3317303} {\path{doi:10.1109/TBIOM.2023.3317303}}.

\bibitem{Liu2023}
F.~Liu, D.~Chen, F.~Wang, Z.~Li, F.~Xu, Deep learning based single sample face recognition: a survey, Artificial Intelligence Review 56~(3) (2023) 2723--2748.
\newblock \href {https://doi.org/10.1007/s10462-022-10240-2} {\path{doi:10.1007/s10462-022-10240-2}}.

\bibitem{yin2023deep}
Y.~Yin, S.~He, R.~Zhang, H.~Chang, X.~Han, J.~Zhang, Deep learning for iris recognition: A review (2023).
\newblock \href {http://arxiv.org/abs/2303.08514} {\path{arXiv:2303.08514}}.

\bibitem{Saritha2024}
B.~Saritha, M.~A. Laskar, A.~M. Kirupakaran, R.~H. Laskar, M.~Choudhury, N.~Shome, Deep learning-based end-to-end speaker identification using time-frequency representation of speech signal, Circuits, Systems, and Signal Processing 43~(3) (2024) 1839--1861.
\newblock \href {https://doi.org/10.1007/s00034-023-02542-9} {\path{doi:10.1007/s00034-023-02542-9}}.

\bibitem{10217750}
Y.~Mohamed, A.~M. Anter, A.~B. Zaky, Recurrent neural networks (rnns) to improve eeg-based person identification, in: 2023 Intelligent Methods, Systems, and Applications (IMSA), 2023, pp. 616--621.
\newblock \href {https://doi.org/10.1109/IMSA58542.2023.10217750} {\path{doi:10.1109/IMSA58542.2023.10217750}}.

\bibitem{he2021grrnn}
S.~He, L.~Schomaker, Gr-rnn: Global-context residual recurrent neural networks for writer identification (2021).
\newblock \href {http://arxiv.org/abs/2104.05036} {\path{arXiv:2104.05036}}.

\bibitem{hybird}
S.~Saleem, A.~Mohsin~Abdulazeez, Hybrid trainable system for writer identification of arabic handwriting, Computers, Materials and Continua 68 (05 2021).
\newblock \href {https://doi.org/10.32604/cmc.2021.016342} {\path{doi:10.32604/cmc.2021.016342}}.

\bibitem{lamisidentification}
S.~Lazrak, A.~Semma, A.~E.~K. Noureddine, Y.~Elkettani, D.~Mentagui, Writer identification using textural features, ITM Web of Conferences 43 (2022) 01027.
\newblock \href {https://doi.org/10.1051/itmconf/20224301027} {\path{doi:10.1051/itmconf/20224301027}}.

\bibitem{10.1371/journal.pone.0284680}
B.~Q. Ahmed, Y.~F. Hassan, A.~S. Elsayed, Offline text-independent writer identification using a codebook with structural features, PLOS ONE 18~(4) (2023) 1--31.
\newblock \href {https://doi.org/10.1371/journal.pone.0284680} {\path{doi:10.1371/journal.pone.0284680}}.

\bibitem{he2015deep}
K.~He, X.~Zhang, S.~Ren, J.~Sun, Deep residual learning for image recognition (2015).
\newblock \href {http://arxiv.org/abs/1512.03385} {\path{arXiv:1512.03385}}.

\bibitem{sandler2019mobilenetv2}
M.~Sandler, A.~Howard, M.~Zhu, A.~Zhmoginov, L.-C. Chen, Mobilenetv2: Inverted residuals and linear bottlenecks (2019).
\newblock \href {http://arxiv.org/abs/1801.04381} {\path{arXiv:1801.04381}}.

\bibitem{tan2020efficientnet}
M.~Tan, Q.~V. Le, Efficientnet: Rethinking model scaling for convolutional neural networks (2020).
\newblock \href {http://arxiv.org/abs/1905.11946} {\path{arXiv:1905.11946}}.

\bibitem{Khatt}
S.~Mahmoud, I.~Ahmad, W.~Al-Khatib, M.~Alshayeb, M.~Parvez, V.~Märgner, G.~Fink, Khatt: An open arabic offline handwritten text database, Pattern Recognition 47 (2014) 1096–1112.
\newblock \href {https://doi.org/10.1109/ICFHR.2012.224} {\path{doi:10.1109/ICFHR.2012.224}}.

\bibitem{Khan2022}
M.~Khan, Arabic handwritten alphabets, words and paragraphs per user (ahawp) (2022).
\newblock \href {https://doi.org/10.17632/2h76672znt.2} {\path{doi:10.17632/2h76672znt.2}}.

\bibitem{djeddi2015lamis}
C.~Djeddi, A.~Gattal, L.~Souici-Meslati, I.~Siddiqi, Y.~Chibani, H.~El~Abed, Lamis-mshd: A multi-script offline handwriting database, in: 2015 13th International Conference on Document Analysis and Recognition (ICDAR), IEEE, 2015, pp. 1146--1150.

\bibitem{Otsu1979ATS}
N.~Otsu, A threshold selection method from gray-level histograms, IEEE Trans. Syst. Man Cybern. 9 (1979) 62--66.

\bibitem{dilation}
T.~Zhang, C.~Suen, A fast parallel algorithm for thinning digital patterns, Commun. ACM 27 (1984) 236--239.

\bibitem{4767851}
J.~Canny, A computational approach to edge detection, IEEE Transactions on Pattern Analysis and Machine Intelligence PAMI-8~(6) (1986) 679--698.
\newblock \href {https://doi.org/10.1109/TPAMI.1986.4767851} {\path{doi:10.1109/TPAMI.1986.4767851}}.

\bibitem{Roi}
R.~Lienhart, J.~Maydt, An extended set of haar-like features for rapid object detection, Vol.~1, 2002, pp. I--900.
\newblock \href {https://doi.org/10.1109/ICIP.2002.1038171} {\path{doi:10.1109/ICIP.2002.1038171}}.

\end{thebibliography}

\end{document}